\documentclass{llncs}

\usepackage[center]{caption}

\usepackage{subcaption}
\usepackage{booktabs}

\usepackage{makeidx}  

\usepackage{lipsum}
\usepackage{url}
\usepackage{hyperref}
\makeatletter \newif\if@restonecol \makeatother  
\usepackage[linesnumbered, ruled, vlined]{algorithm2e}
\usepackage{algpseudocode}

\usepackage[bottom]{footmisc}

\usepackage{float}
\usepackage{epsfig}
\usepackage{epstopdf}
\usepackage{graphics}
\DeclareGraphicsExtensions{.eps,.gz,.eps,.pdf,.png,.jpg}

\usepackage{listings}
\usepackage{graphicx}
\usepackage{amsmath,amssymb,amsfonts}
\usepackage{enumerate}
\usepackage{paralist}
\usepackage{multirow}
\usepackage{verbatim}
\usepackage{dblfloatfix}
\usepackage[table]{xcolor}
\usepackage[export]{adjustbox}
\usepackage{threeparttable}
\usepackage{tabularx}

\newcommand{\comments}[1]{}
\newcommand{\ignore}[1]{}
\newcommand{\eat}[1]{}

\newcommand{\sstitle}[1]{\smallskip\noindent{\bf #1.\/}}

\newcolumntype{L}[1]{>{\raggedright\let\newline\\\arraybackslash\hspace{0pt}}m{#1}}
\newcolumntype{C}[1]{>{\centering\let\newline\\\arraybackslash\hspace{0pt}}m{#1}}
\newcolumntype{R}[1]{>{\raggedleft\let\newline\\\arraybackslash\hspace{0pt}}m{#1}}

\makeatletter
\newcommand{\removelatexerror}{\let\@latex@error\@gobble}
\makeatother

\usepackage{pslatex}

\begin{document}


\setlength{\belowdisplayskip}{3pt}
\setlength{\belowdisplayshortskip}{3pt}
\setlength{\abovedisplayskip}{3pt}
\setlength{\abovedisplayshortskip}{3pt}

\title{Multimodal Classification for Analysing Social Media}

  \author{%
  {
  Chi Thang Duong \and
  Remi Lebret \and
  Karl Aberer
  }%
  }

\institute{\'{E}cole Polytechnique F\'{e}d\'{e}rale de Lausanne \\
\email{\{thang.duong,remi.lebret,karl.aberer\}@epfl.ch}
}

\maketitle


\begin{abstract}
Classification of social media data is an important approach in understanding user behavior on the Web. Although information on social media  can be of different modalities such as texts, images, audio or videos, traditional approaches in classification usually leverage only one prominent modality. Techniques that are able to leverage multiple modalities are often complex and susceptible to the absence of some modalities. In this paper, we present simple models that combine information from different modalities to classify social media content and are able to handle the above problems with existing techniques. Our models combine information from different modalities using a pooling layer and an auxiliary learning task is used to learn a common feature space. We demonstrate the performance of our models and their robustness to the missing of some modalities in the emotion classification domain. Our approaches, although being simple, can not only achieve significantly higher accuracies than traditional fusion approaches but also have comparable results when only one modality is available.
\end{abstract}

\section{Introduction}
\label{sec:intro}
With the advance of social media networks, people are sharing user-generated contents in an unprecedented scale~\cite{retain2017}. Social media networks have shifted from being specialized as users can only share either texts or images to general-purpose as users can now share texts, images, audio segments or even video clips. As recent statistics show that posts with images get more interaction~\footnote{\href{https://blog.bufferapp.com/the-power-of-twitters-new-expanded-images-and-how-to-make-the-most-of-it/}{blog.bufferapp.com/the-power-of-twitters-new-expanded-images[..]}} while video tweets fuels discovery and drives engagement~\footnote{\href{https://blog.twitter.com/2015/new-research-twitter-users-love-to-watch-discover-and-engage-with-video}{blog.twitter.com/2015/new-research-twitter-users[..]}}, we expect an enormous increase of \emph{multimodal posts}. The contents posted by users on social networks are a way for them to express themselves such as their emotions, feelings or to share life events. As a result, these posts provide precious information about the users that if these information is analyzed systematically, we would have a strong understanding about the users. For instance, knowing the emotions of the users by analyzing these contents can bring enormous benefits. An ad campaign can customize its ad based on the emotion expressed in a recent post by a user, which would reduce marketing cost but increase user engagement. A further step is to induce an emotion in a user proactively using a specific image or piece of text. A use case would be using happy images or joyful texts to encourage users to buy a product.
There are many approaches to social media analysis and classification is usually the tool of choice when we want to study an aspect of the user base in depth such as their sentiment or emotion.

\begin{table}[t]
\centering
\begin{tabular}{c|c|c}
\includegraphics[scale=0.2,valign=c]{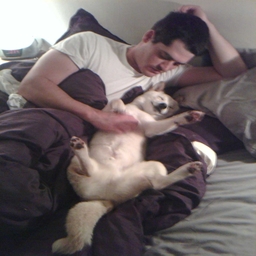} & \includegraphics[scale=0.2,valign=c]{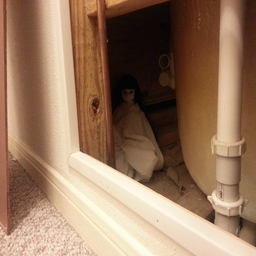} & \includegraphics[scale=0.2,valign=c]{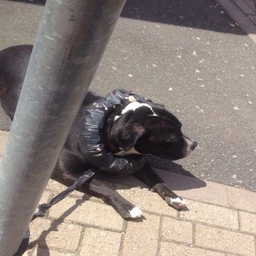} \\
\hline
\parbox[t]{2cm}{This is my happy.} & \parbox[t]{2cm}{[...] I decided to leave a little present for the next tenants.} & \parbox[t]{2cm}{Poor dog was wearing [...] bricks [...] to be a fighting dog :(} \\
\end{tabular}
\caption{Example posts on social media with both image and image descriptor}
\label{tbl:example}
\vspace*{-20pt}
\end{table}


Although classifying social media content for user understanding is highly important, traditional works tend to focus on one modality\cite{cui2010multiple}. For instance, works in emotion classification leverage only the visual information, textual information, which is also important, is often left out. Although images can express users' emotions more vividly, textual information can also convey various emotional aspects. Table \ref{tbl:example} shows some posts with the availability of both image and text. In the first example, both the text and the image express the same emotion \emph{joy}, which is also the emotion expressed by the post. On the other hand, in the second example, using only the text, it is easy for us to tell it conveys the emotion \emph{joy} but the visual signal is the dominant one in this case and it expresses the emotion \emph{fear}. In the last example, without considering the text, the image may describe the emotion \emph{contentment} due to the presence of a pet. However, the text provides the context which changes the meaning and the emotion expressed in the image completely.
These examples have shown that when information is available in different modalities, only by leveraging all of them can give us a complete picture.


Given the drawbacks of unimodal approaches, several works have been done on combining multimodal information for social media analysis. These works are able to improve the classification accuracy significantly\cite{zeppelzauer2016multimodal}. However, most of these approaches are usually complicated and tailored to specific problems and modalities, which makes them hard to apply to new problems or add other modalities\cite{cui2010multiple}. In addition, these classification techniques also assume the availability of all the required modalities, which is not the case in practice. For instance, a tweet classification technique that combines visual and textual information to classify may not work if the tweets contain only texts. Inspired by these observations, we propose techniques to combine information from multiple modalities based on neural network models. Our models combine information from different modalities using a pooling layer and an auxiliary learning task is used to learn a common feature space for all modalities. Our techniques are able to achieve high accuracy without the use of complex model while allowing to classify even when some modalities are missing. In addition, our techniques are easily scalable to new problems or more modalities.

We demonstrate the performance of our approaches in the emotion classification domain. Emotion classification is a multiclass classification problem with many applications in practice. We focus on two prominent modalities which are textual and visual resources.  To the best of our knowledge, we are the first to propose techniques that combine information from different modalities in the emotion classification domain. 
In this paper, we make the following contributions.

\begin{itemize}
  \item We propose novel generic techniques that can combine information from multiple modalities to classify social media content. In particular, we leverage advances in neural networks to build our fusion approaches that are scalable to new modalities and new problems.
  \item Our proposed techniques are also robust to different types of input as they are able to handle the missing of some modalities. In the emotion classification demonstration, they can tackle three cases: only image or text or both are available. 
  \item We construct a dataset which contains data from textual and visual modalities to test our techniques. The dataset also contains strong labels for training and testing. We also enrich an image-only dataset with textual data. These datasets will be made available to foster research in the emotion analysis field.

\end{itemize}

The rest of the paper is organized as follows. Section \ref{sec:related_work} introduces related works on the field of multimodal classification and emotion analysis. Section \ref{sec:model} discusses a general fusion model and traditional fusion approaches. Section \ref{sec:joint} explains in detail our proposed techniques. Experimental evaluation and analysis are presented in Section \ref{sec:experiments} while Section \ref{sec:conclusion} concludes the paper.

\section{Related Work}
\label{sec:related_work}

\sstitle{Multimodal classification for social media} Multimodal classification techniques can be classified into two main classes depending on how the information from multiple modalities are combined. In \emph{late fusion}, separate classification results obtained from different modalities are constructed first and the fusion is done based on these results at the decision level\cite{cui2010multiple}. Late fusion implies an assumption that the underlying data from different modalities are independent. This assumption is not practical as information from different modalities still describe the same event/object, which means they could be correlated\cite{sargin2007audiovisual}. In a recent work\cite{you2016cross}, a variant of late fusion is discussed where the authors used KL divergence to enforce the results from different modalities to be similar. \emph{Early fusion} takes a different approach as the fusion is done at the feature level where information from different modalities are appended. Classification is then done on the appended representation\cite{zeppelzauer2016multimodal}. Various variants of early fusion are proposed to classify social media content. In sentiment analysis, instead of concatenating information from different modalities, the work in \cite{you2016robust} use LSTM to combine visual and textual information. One approach\cite{zeppelzauer2016multimodal} in social event classification constructed a hierarchical classifier on the concatenated features to classify non-event and other event types. Several intermediate fusion techniques are also available such as using LDA to extract joint latent topics\cite{barnard2003matching} or by using statistical methods such as CCA\cite{kim2007discriminative} for image classification. Existing techniques for multimodal classification are usually complex and they often require the presence of all modalities. Although our proposed approaches can be considered as a variant of early fusion, they differ from previous techniques as they can handle the absence of some modalities while being simpler. Our model seems to be similar to a Siamese network~\cite{koch2015siamese} which contains two identical subnetworks with the same parameters and weights. However, our approach differs from a Siamese network as the image and text subnetworks’ architecture are different, hence, their parameters are completely different.

\sstitle{Emotion analysis} Techniques for analyzing emotions can be classified into hand-crafted features and deep features. Early works in emotion analysis leverage features based on art and psychology theory. These artistic features are usually low-level such as shape~\cite{lu2012shape}, color and texture~\cite{machajdik2010affective}. Several higher-level features which are the combination of these low-level features are also used~\cite{irie2010affective,jia2012can,borth2013sentibank,argmine2017}. In a recent work~\cite{zhao2014exploring}, the authors proposed several features based on principles of arts in which combinations of these features can evoke different emotions. After these features are defined, a classifier such as SVM is constructed on top of these features to classify the emotions. The inherent drawback of techniques that uses hand-crafted features is the required availability of these features. However, designing these features is a tedious process which requires expert knowledge and it is not guaranteed that all features that can capture the emotions are covered. Given this drawback, techniques that do not require a feature design process are proposed. These features are constructed automatically using CNNs. Only a recent work\cite{chen2015learning} has leveraged deep features to emotion analysis and already achieved better accuracy in comparison with hand-crafted features. However, the technique in \cite{chen2015learning} only deals with images, which leaves out the textual information completely. As we show in Table\ref{tbl:example}, considering only textual or visual information is not enough to analyze emotions. This work is also the closest to our work as we also use build image representations using Convolutional Neural Networks(CNNs). However, we differ from this work significantly as we also take into account textual information and the combination of them. To the best of our knowledge, our technique is the first to combine visual and textual information to classify emotions.

\section{Multimodal Classification for Analysing Social Media}
\label{sec:model}

In this section, we denote the modalities in social media that we use in our models,
and we present the traditional approaches for combining them in classification.

\subsection{Multimodality in Social Media}

Multimodality describes communication practices in terms of the textual, aural,
linguistic, spatial, and visual resources - or modes - used to compose messages
(or \emph{posts}) \cite{lutkewitte2013multimodal}.
In this paper, we focus on combining the two prominent modalities of social media,
i.e. textual and visual resources. It is worth mentioning that the proposed models
is easily scalable to more modalities.

We define $x \in X$ a \emph{post} that can be composed of an \emph{image} or a \emph{text},
or both an image and a text.
When both image and text are present, we assume that they are semantically related,
e.g. the text describes the image.
Each image $i$ is represented with an image feature vector $\gamma(i) \in \mathbb{R}^{n}$.
Thanks to the recent progress in computer vision using deep learning, $\gamma(i)$
can be extracted from CNNs trained on millions of images.
A text can be as long as a paragraph or as short as a phrase.
We represent each piece of text $s$ with a textual feature vector $\psi(s) \in \mathbb{R}^{m}$.
Such textual feature vectors can be obtained with classical bag-of-words models
or with approaches based on word embeddings.

\subsection{Multimodal Classification using \\Feature Level Fusion}

\begin{figure*}[tbh!]
  \centering
     \begin{subfigure}{.23\linewidth}
  \centering
 \includegraphics[width=0.9\linewidth]{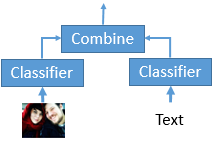}
 \caption{Late fusion}
 \label{fig:late}
  \end{subfigure}
  \quad
   \begin{subfigure}{.23\linewidth}
  \centering
 \includegraphics[width=1.0\linewidth]{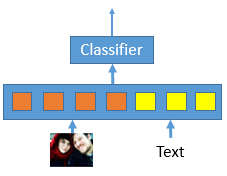}
    \caption{Early fusion}
      \label{fig:early}
  \end{subfigure}
  \quad
 \begin{subfigure}{.2\linewidth}
\centering
 \includegraphics[width=1.0\linewidth]{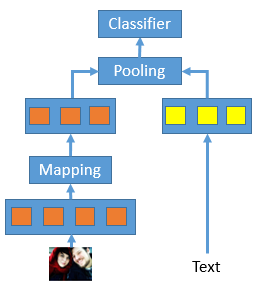}
    \caption{Joint fusion}
      \label{fig:joint}
  \end{subfigure}
  \quad
 \begin{subfigure}{.2\linewidth}
\centering
 \includegraphics[width=1.0\linewidth]{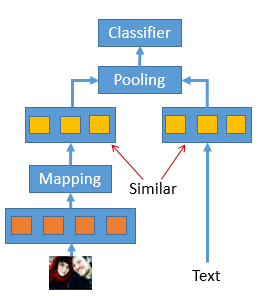}
    \caption{Common space fusion}
      \label{fig:common}
  \end{subfigure}
  \caption{Traditional techniques (a,b) and our techniques (c,d)}
  \label{fig:models}
  \vspace*{-15pt}
\end{figure*}

Given a post $x \in X$ and a set of classes $Y$, we define a model that assigns
probabilities to all $y \in Y$.
The predicted class is then the one with the highest probability,
\begin{equation}\label{eq:prob}
  \hat{y} = \arg\max_{y} P(Y = y | x).
\end{equation}
When $x=i$ or $x=s$, unimodal classifiers are trained on the entire training set
in each modality.
As we are interested in combining image and text (i.e. when $x=\{i,s\}$), we discuss in this part
two traditional fusion approaches: \emph{late} fusion and \emph{early} fusion.
The difference in these approaches is when the fusion
is done in the classification process. In early fusion, the fusion happens before
the classifier is constructed (hence the name early) while in late fusion,
the fusion is done after the classifiers are created.

\subsubsection{Late fusion}

Late fusion requires the construction of two independent classifiers: one for image
and one for text. The predicted class is then the highest product between the two classifiers,
\begin{equation}\label{eq:late}
  \hat{y} = \arg\max_{y} P(Y = y | i)P(Y = y | s).
\end{equation}

\subsubsection{Early fusion}
Early fusion, on the other hand, does not require the construction of two separate classifiers as the fusion applies in the feature space. More precisely, it requires modeling the post $x$ as a feature vector:
\begin{equation}\label{eq:early}
\mathbf{x} = [\gamma(i); \psi(s)]
\end{equation}
where $\mathbf{x} \in \mathbb{R}^{(n+m)}$ is the vector for the post $x$ and $[\mathbf{v}_1;\mathbf{v}_2]$ denotes vector concatenation. The model to classify $x$ can be constructed on top of the feature vectors $\mathbf{x}$. Classification techniques such as SVM can be used. In our work, the classification is done using neural network as discussed in the next section.

\section{Joint Fusion with Neural Network Models}
\label{sec:joint}
Although late fusion and early fusion are able to combine visual and textual
data for classification, they also suffer from some drawbacks.
Late fusion requires the construction of two classifiers. In many cases,
the construction of two separate classifiers would negate the purpose of
combining them as the improvement (if any) from using only one classifier could
be insignificant. On the other hand, while early fusion does not need two
separate classifiers, it requires the availability of both the image and the text,
which may be too stringent. In this section, we propose two fusion approaches
that enjoys the simplicity of early fusion and the flexibility of late fusion.
We call these approaches \emph{joint fusion} and \emph{common space fusion}.
Our fusion approaches are based on neural networks as they provide many advantages: 1) neural networks allow to train the classifier in an end-to-end manner without involving the tedious feature engineering process, 2) neural networks are also highly-customizable as the model can be changed easily by adding or removing some layers.

\subsection{Mathematical notations and layers}

A feedforward neural network estimates $P(Y = y | x)$ with a parametric function
$\phi_\theta$ (Equation~\ref{eq:prob}), where $\theta$ refers to all learnable parameters
of the network.
Given an input $x$, this function $\phi_\theta$ applies a combination of functions
such as
\begin{equation}
  \phi_\theta(x) = \phi^{L}(\phi^{L-1}(\ldots\phi^1(x)\ldots)),
\end{equation}
with $L$ the total number of layers in the network.

We denote matrices as bold upper case letters ($\mathbf{X}$, $\mathbf{Y}$, $\mathbf{Z}$),
and vectors  as bold lower-case letters ($\mathbf{a}$, $\mathbf{b}$, $\mathbf{c}$).
$\mathbf{A}_i$ represents the $i^{\text{th}}$ row of matrix $\mathbf{A}$ and $[\mathbf{a}]_i$ denotes the $i^{\text{th}}$ element of vector $\mathbf{a}$.
Unless otherwise stated, vectors are assumed to be column vectors.
We now introduce the two standard layers when training linear classifiers
with neural networks: the \emph{linear} layer and the \emph{softmax} layer.

\subsubsection{Linear layer}
This layer applies a linear transformation to its inputs $\mathbf{x}$:
\begin{equation}
  \boldsymbol{\phi}^l(\mathbf{x}) = \mathbf{W}^l\mathbf{x} + \mathbf{b}^l\,
\end{equation}
where $\mathbf{W}^l$ and $\mathbf{b}^l$ are the trainable parameters with
$\mathbf{W}^l$ being the weight matrix, and $\mathbf{b}^l$ is the bias term. For instance, when training an image classifier, $\mathbf{x} = \gamma(i) \in \mathbb{R}^n$
with $\mathbf{W}^l \in \mathbb{R}^{|Y| \times n}$ and $\mathbf{b}^l \in \mathbb{R}^{n}$.
And $\mathbf{x} = \psi(s) \in \mathbb{R}^m$ with $\mathbf{W}^l \in \mathbb{R}^{|Y| \times m}$
and $\mathbf{b}^l \in \mathbb{R}^{m}$, when training a classifier with only textual content. 

\subsubsection{Softmax layer}
Given an input $x$, the penultimate layer outputs a score for each class $y \in Y$,
$\boldsymbol{\phi}^{L-1}(x) \in \mathbb{R}^{|Y|}$.
The probability distribution is obtained by applying the softmax activation function:
\begin{equation}
P(Y = y | x) \propto \phi_\theta(x,y) = \frac{\exp(\boldsymbol{\phi}^{L-1}(x,y))}{ \sum_{k=1}^{|Y|} \exp(\boldsymbol{\phi}^{L-1}(x,y_k))}
\end{equation}

Using the above layers, early fusion can be described as a neural network as in the following example.
\begin{example}[Early fusion as a neural network model]
Early fusion can be modelled as a neural network with 3 layers: $\phi_\theta(x) = \phi^{3}(\phi^{2}(\phi^1(x)))$. The first layer $\phi^1$ is a \emph{fusion layer} that applies the concatenation operation on the image and text feature vectors, $\gamma(i),\psi(s)$. The output of this layer is the post vector $\mathbf{x} = [\gamma(i); \psi(s)]$. The second layer $\phi^2$ is a linear layer that takes the concatenation of both image and text feature vectors, $\mathbf{x} = [\gamma(i); \psi(s)] \in \mathbb{R}^{(n+m)}$. The learnable parameters of this layer are $\mathbf{W}^2 \in \mathbb{R}^{|Y| \times (n+m)}$ and $\mathbf{b}^2 \in \mathbb{R}^{(n+m)}$. The final layer $\phi^3$ is a softmax layer which converts the score for each class obtained from the second layer to probability. The parameters of early fusion are $\theta = \{\mathbf{W}^2, \mathbf{b}^2\}$

\end{example}

\subsection{Feature Vectors}
Before diving into the detail of our fusion models, we discuss the process to represent each image and text as feature vectors ($\gamma(i), \psi(s)$) as \emph{early fusion} and our proposed approaches require an underlying image and text representation. A good image and text representation can affect the performance of these approaches heavily.

\subsubsection{Image Representations}
To represent the images, we use a CNN to extract
features, i.e. represent each image $i$ as a feature vector $\gamma(i) \in \mathbb{R}^n$. This is motivated by
the fact that CNNs are able to obtain state-of-the-art results in many object
classification tasks\cite{ciresan2011flexible}. Therefore, the CNN may capture
features which are suitable to classify objects\cite{sharif2014cnn}.
These features may also be applicable for other types of classification, such as emotion.
Technically, we take a CNN\cite{szegedy2015going} trained on the ImageNet dataset
for the object classification task and remove the last fully-connected layer
while keeping the other layers the same. By feeding each image through this
pretrained CNN, we obtain an image vector in the output.

It is worth noting that by keeping other layers the same, we use no parameter to obtain the image vector. However, we can also retrain the CNN with our dataset. In this case, the parameter for $\gamma(i)$ is all the parameters of the CNN. They will be trained together with other parameters of the joint fusion models.

\subsubsection{Text Representations}
The process of converting a text $s$ from its original format (i.e. words)
to a $d$-dimensional vector is modelled by a function $\psi_{\mu}(s)$, which takes
a sequence of words $s = \langle w_1,w_2,...,w_n \rangle$ as input,
where each word $w_t$ comes from a predefined vocabulary $\mathcal{V}$.
This function is a composition of an \emph{embedding} layer and an
\emph{aggregation} layer.

\sstitle{Embedding layer}
Given a parameter matrix $\mathbf{E} \in \mathbb{R}^{|\mathcal{V}| \times d}$,
the \emph{embedding layer} is a lookup table that performs an array indexing operation:
\begin{equation}
\psi^1(w_t) = \mathbf{E}_t \in \mathbb{R}^d,
\end{equation}
where $\mathbf{E}_{t}$ corresponds to the embedding of the element $w_i$ at row $i$.
This matrix is usually initialized with pretrained word embeddings.
$d$ is the dimensionality of the word vectors which is a hyperparameter chosen by the users.

By feeding each word of a text $s = \langle w_1,w_2,...,w_n \rangle$ through the lookup table,
we obtain the following output matrix:
\begin{equation}
\psi^1(s) = ( \mathbf{E}_{1},\mathbf{E}_{2},...,\mathbf{E}_{n} ) \in \mathbb{R}^{d \times n}.
\end{equation}

\sstitle{Aggregation layer}
The second layer $\psi^2$ of the network is an aggregation layer that takes
the matrix $\psi^1$ from the previous layer as input and returns
a vector representation for the text $s$.
Formally, the aggregation layer applies the following operation:
\begin{equation}
\psi^2(s) =  \text{agg} \big\{ \boldsymbol{\psi}^1(w_t), \forall w_t \in s \big\},
\end{equation}
where $agg$ is an aggregation function. This function can be either the average
or a component-wise max.
More complex functions could be used for aggregating the word embeddings, such as
a convolutional or recurrent neural network. But it has been shown that these simpler
aggregating function with no parameters have similar performance on classification tasks~\cite{joulin2016bag}.

As the network $\psi_{\mu}$ has only two layers and the second layer has no parameter,
$\mu=\{\mathbf{E}\}$. This parameter will be trained together with the parameters of the classifier.

\subsection{Joint Fusion Models}
As discussed above, our motivation for joint fusion is to have models that can classify posts having only image or text or both without constructing two classifiers like late fusion. To achieve this goal, our joint fusion models change how the post vector $\mathbf{x}$ is constructed from the image and text vector.

\subsubsection{Joint Fusion Learning}
In early fusion, the post vector $\mathbf{x}$ is constructed in the fusion layer by concatenating the image vector $\gamma(i)$ and the text vector $\psi(s)$. This way of combining the vectors has two important implications. First, both the image and text vector must be available, which prevents our intention to classify using only image or text. Second, we still consider the textual and visual information as separate as each of them affects the classification independently. In joint fusion models, we change the way the text and image vectors are fused in the fusion layer while keeping other layers similar to early fusion.

\sstitle{Fusion layer}
The fusion layer takes the image vector and text vector $\gamma(i), \psi(s)$ as input and applies the pooling operation to obtain the post vector $\mathbf{x}$:
\begin{equation}
\mathbf{x} = \text{pooling}(\gamma(i),\psi(s))
\end{equation}
The pooling function can be either a \emph{component-wise max} pooling, or an \emph{average}
pooling.

It is worth noting that the pooling operation requires the vectors $\gamma(i) \in \mathbb{R}^n$
and $\psi(s) \in \mathbb{R}^m$ to have the same size. In Figure \ref{fig:joint}, we assume the image vector has higher dimension and we project its feature vector into the textual feature space.
This can be done by adding an extra linear layer
to $\gamma$ (i.e. the network that extracts image feature vector). Assuming $n > m$, the linear layer is as follows:
\begin{equation}
\tilde{\gamma}(i) = \tilde{\mathbf{W}}\gamma(i) + \tilde{\mathbf{b}}
\end{equation}
where $\tilde{\gamma}(i) \in \mathbb{R}^m$, $\tilde{\mathbf{W}} \in \mathbb{R}^{n\times m}$ and $ \tilde{\mathbf{b}} \in \mathbb{R}^m$. The input to the fusion layer is then two vectors $\ddot{\gamma}(i)$ and $\psi(s)$.

The parameters of joint fusion are $\theta = \{\mathbf{W}^2, \mathbf{b}^2, \tilde{\mathbf{W}}, \tilde{\mathbf{b}}\}$.

\sstitle{Training} The parameter $\theta$ is obtained by training the joint fusion neural network by minimizing the negative log-likelihood using stochastic gradient descent(SGD):
\begin{equation}
L(\theta) =  \sum_{(x,y)}-\log P(Y=y | x) \propto \sum_{(x,y)} - \log \big(\phi_\theta(x,y)\big).
\end{equation}

\subsubsection{Common Feature Space Fusion} 
Although joint fusion allows to classify even when only image or text is available, the accuracy of these unimodal classifiers are not the same (c.f. Section~\ref{sec:experiments}) as joint fusion still considers visual and textual signals of the same post as different. Motivated by this observation, we propose \emph{common space fusion} which aims to enforce visual and textual vectors of the same post to be similar i.e. to be in the same feature space. We achieve this using an \emph{auxiliary learning task} in addition to the classification main task. The neural network of common space fusion is similar to that of joint fusion with the exception of the addition of the auxiliary task.

\sstitle{Auxiliary learning task} The goal of the auxiliary learning task is to make the image and text vector $\gamma(i),\psi(s)$ of the post $x$ to be similar while the image vector $\gamma(i)$ of the post $x$ and the text vectors $\{\psi(s^-)\}$ of posts from different classes are different. The vectors $\langle \gamma(i),\psi(s^+) \rangle$ of the post $x$ is called a \emph{positive} pair while the vectors $\langle \gamma(i),\psi(s^-) \rangle$ is called a \emph{negative} pair. We measure the similarity and difference between the pairs using a distance metric $d(\gamma(i),\psi(s))$. Intuitively, we want the distance of the positive pair to be low while the distance of the negative pairs to be high. This objective is captured by the loss function for the auxiliary task.

\sstitle{Training} Traditionally, the objective of the auxiliary task is captured using a margin-based loss function. However, a recent work has shown that using a probabilistic interpretation of the margin-based loss function yields better result\cite{hoffer2016deep}. The loss function for the auxiliary task is defined as follows:
\begin{multline}
L(i,s^+,\{s^-_j\}_{j=1,g}) = -\sum_{j=1,g} \log\Big( \frac{\exp(-d(\gamma(i),\psi(s^+)))}{\exp(-d(\gamma(i),\psi(s^+))) + \exp(-d(\gamma(i),\psi(s^-_j)))}\Big)
\end{multline}
where $g$ is the number of negative pairs to be used in training.
With $\theta = \{\mathbf{W}^2, \mathbf{b}^2, \tilde{\mathbf{W}}, \tilde{\mathbf{b}}\}$, we minimize the following loss function involving the auxiliary learning task and the classification main task for a given training sample $(x,y)$, where $x=\{i,s^+\}$:

\begin{multline}
L(x, y; \theta) = -\lambda \log\big(\phi_\theta(x,y)\big) -\sum_{j=1,g} \log\Big( \frac{\exp(-d(\gamma(i),\psi(s^+)))}{\exp(-d(\gamma(i),\psi(s^+))) + \exp(-d(\gamma(i),\psi(s^-_j)))}\Big)
\end{multline}
where the $g$ negative text samples $s^-$ are chosen randomly at each iteration, and $\lambda$ is a hyperparameter specifying the weight of the classification main task.

\section{Experiments with Emotion Classification}
\label{sec:experiments}
In this section, we evaluate our proposed fusion approaches on the emotion classification application. To the best of our knowledge, we are the first one to study combining visual and textual data for emotion classification.

\subsection{Emotion as Discrete Categories}

\begin{figure}[h]
\vspace*{-15pt}
  \centering
  \includegraphics[width=0.6\linewidth]{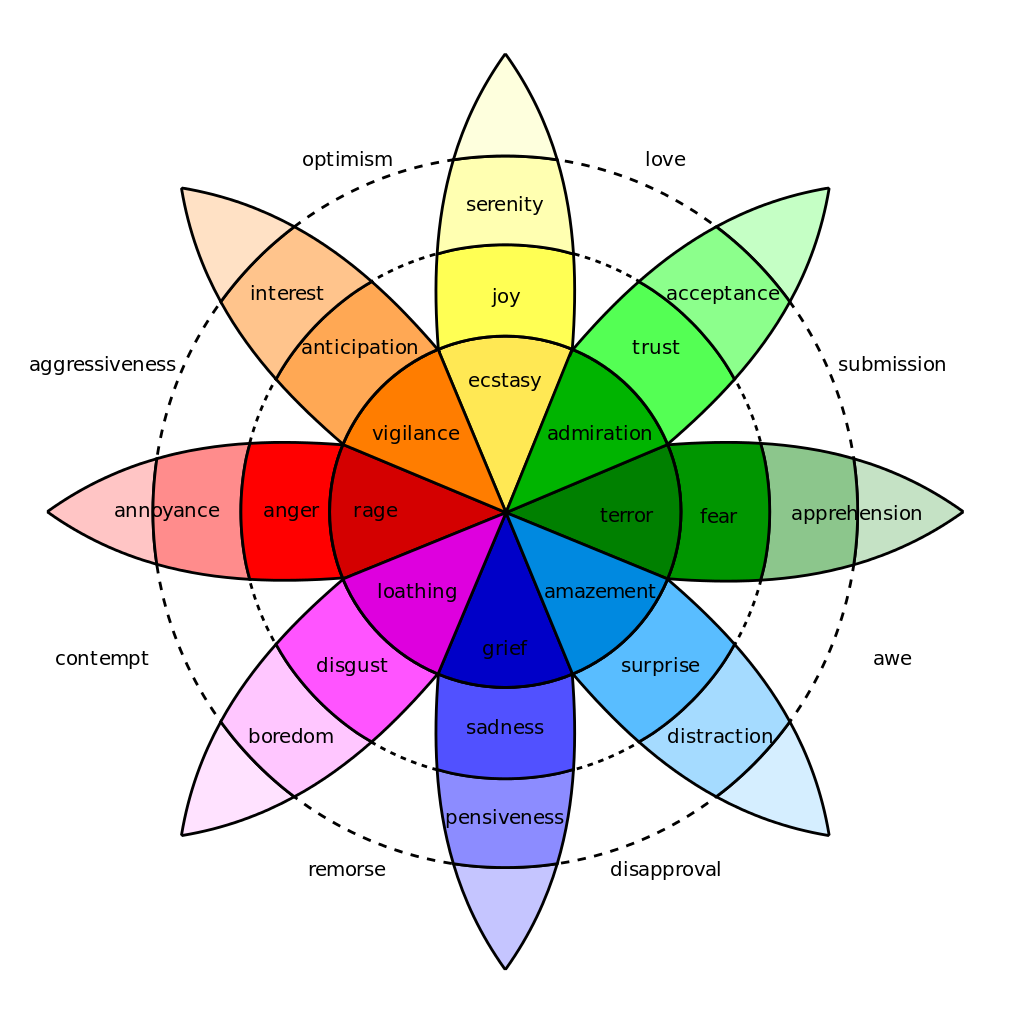}\\
  \vspace*{-15pt}
  \caption{Plutchik's classification of emotions}\label{fig:plutchik}
  \vspace*{-10pt}
\end{figure}

In emotion classification, the problem we want to solve is given a post $x$, we want to construct a classifier
to find out the post's membership i.e., $x$ belongs to which emotion class.
The set of emotion classes is denoted as $Y=\{y_1,...,y_k\}$ where each class $y_i \in Y$
and $\forall i,j: y_i \cap y_j = \emptyset $.
For the emotion classes, we use the most well-known Plutchik's classification of emotions
\cite{plutchik2001nature}. An illustration of the Plutchik's wheel of emotions
is shown in Figure \ref{fig:plutchik}.

\subsection{Datasets}
\noindent
To the best of our knowledge, there is no large-scale dataset for emotion classification that contains both visual and textual data. Even for visual content only, most of the datasets are relatively small except for the \emph{flickr} dataset\cite{you2016building}. This motivates us to build a dataset from scratch while also adding textual data to the \emph{flickr} dataset.

\sstitle{Enriching an image-only dataset} The \emph{flickr} dataset was produced by You et. al.\cite{you2016building}, which contains images from a popular image sharing website\footnote{www.flickr.com}. To obtain the labels for these images, at least 5 workers from Amazon Mechanical Turks were hired to label them into 8 emotion classes: \emph{amusement, anger, awe, contentment, disgust, fear} and \emph{sadness}. We first crawl the images from this dataset while keeping only the ones where the majority of the workers agree that the image belongs to a specific emotion class. Then, for each image, we collect its title and description to use as textual data. Although these information are not provided in the dataset, they are still available from the image sharing website. We only keep the images with English title and description and the total of words in the title and description is more than 5. The statistics of the \emph{flickr} dataset is shown in Table \ref{tbl:datasets}.

\sstitle{Building an emotion dataset from scratch} In order to test our approaches in different settings, we build another emotion dataset by crawling data from Reddit\footnote{www.reddit.com}. Reddit is a discussion website where discussions are organized by topics (i.e. subreddits). Reddit also has a reputation system where submissions are vote up or down. This reputation system enforces that 1) submissions in a subreddit are always belong to a topic, 2) each submission is ``labelled'' by a number of users. These characteristics make Reddit an attractive social media to build an emotion dataset as it provides strongly-labelled data while a submission always contains text in its title and sometimes contains image. We focus on 4 popular subreddits (\emph{happy}, \emph{creepy}, \emph{rage}, \emph{gore}) which are related to emotion and contain high amount of image submissions. These subreddits correspond to the emotion \emph{joy}, \emph{fear}, \emph{anger}, \emph{disgust} in the Plutchik's model for emotion classification.

For each subreddit, we crawled 1000 submissions with the highest number of upvotes. We kept only posts containing image and discarded the rest. This created an imbalance in the number of posts between the emotion classes. For this reason, we collected submissions with at least 100 upvotes for two classes with the least amount of posts. It is worth noting that one post may contain several images. All the images are then converted to jpg format. The statistics of the \emph{reddit} dataset is shown in Table \ref{tbl:datasets}.

\begin{table}[t]
\begin{minipage}[t]{.54\linewidth}
\centering
\caption{Statistics of the two datasets}
\label{tbl:datasets}
\footnotesize
\begin{tabular}[t]{cll}
\toprule
\multicolumn{1}{l}{} &\textbf{emotion}      &\textbf{\#posts}      \\ \cmidrule(r){1-3}
\multirow{4}{*}{reddit}       & joy       & 1119 \\ 
                              & fear      & 697  \\ 
                              & anger        & 613  \\ 
                              & disgust        & 810  \\ 
                              \cmidrule(r){2-3}
                              & \textbf{Total}       & 3239 \\
\bottomrule
\end{tabular}
\begin{tabular}[t]{cll}
\toprule
\multicolumn{1}{l}{} &\textbf{emotion}      &\textbf{\#posts}      \\ \cmidrule(r){1-3}
\multirow{8}{*}{flickr}       & amusement   & 1259 \\ 
                              & anger       & 407  \\ 
                              & awe         & 1561 \\ 
                              & contentment & 2389 \\ 
                              & disgust     & 852  \\ 
                              & excitement  & 1451 \\ 
                              & fear        & 434  \\ 
                              & sadness     & 984  \\ 
                              \cmidrule(r){2-3}
                              & \textbf{Total}       & 9337 \\
                              \bottomrule
\end{tabular}
\end{minipage}
\quad
\begin{minipage}[t]{.4\linewidth}
\centering
\caption{Model hyperparameters}
\label{tbl:para}
\footnotesize
\begin{tabular}{ll}
\toprule
\textbf{Parameter} & \textbf{Value} \\
\cmidrule(r){1-2}
Word embedding size                               & $d = 200$\\
Hidden vector space                                     & $h =  100$ \\
\# negative samples (reddit) & $g = 3$\\
\# negative samples (flickr) & $g=4$ \\
Classification main task weight & $\lambda=3$\\
Aggregation layer function                                   &    max\\
Fusion layer function      &   max\\
\bottomrule
\end{tabular}
\end{minipage}
\vspace*{-20pt}
\end{table}

\subsection{Baselines}
We compare our proposed approaches (joint fusion and common space fusion) with two unimodal baselines (a text-based and an image-based classifier) and two traditional multimodal fusion techniques (early and late fusion). Regarding the text-based classifier, we chose to use fastText\cite{joulin2016bag} which is a shallow network model for text classification. FastText is able to achieve state-of-the-art result without sacrificing much performance in comparison with other deep models\cite{joulin2016bag}. For the image-based classifier, we use a pretrained InceptionNet\cite{szegedy2015going} which is a CNN for object classification task. To make an image classifier for our setting, we replace the last layer in InceptionNet with a linear layer and train this layer with our dataset while keeping other layers the same. In addition to serving as a baseline, we also use InceptionNet to extract the image features as mentioned in Section \ref{sec:joint}, which results in a total of 2048 features per image.
Regarding late fusion, for comparison purpose, we also reuse the text-based and image-based classifiers to obtain the class probabilities. It is worth noting that we have considered other baselines such as using SVM as the classifier. However, as these techniques have worse performance than deep-learning approaches, we do not include them in the paper.

\subsection{Experimental settings}
For the embedding layer, we used the GloVe word vectors\cite{pennington2014glove} trained on Twitter data as they are in the same social media domain as Flickr and Reddit.
For regularization, we used a dropout layer with a dropout probability of 0.25 right after the lookup table to reduce overfitting. 
In addition, it is reported that factorizing the linear classifier into low rank matrices may improve the classification accuracy\cite{mikolov2013efficient}. We also followed this approach by adding a linear layer right before the last layer to map the concatenated vector (in the case of early fusion) and the pooled vector (in the case of joint fusion models) to a hidden vector space with a size of $h$.
Regarding the hyperparameters, we tested different values of them on the validation set and select the ones that gave the best results. Table \ref{tbl:para} describes other hyperparameters. Our models were trained with a learning rate set to 0.01.

The models were trained on a server equipped with a Tesla GPU. For testing, we use the top 10\% posts with the highest number of upvotes for the \emph{reddit} dataset and the highest number of agreements for the \emph{flickr} dataset. The next 10\% of these dataset are used for validation. We use the same splits for all the models in our experiments. All the source codes and the datasets are available at https://emoclassifier.github.io/

\subsection{Quantitative analysis}

\subsubsection{Results on the Reddit dataset}
Table \ref{tbl:exp_reddit} shows the experimental results for the Reddit dataset. The results show that visual and textual classifiers have similar accuracy on the reddit dataset as their difference in accuracy is only 1\%. In addition, all the fusion techniques have better accuracy than the baselines. The traditional fusion approaches improve the accuracy by at least 5\%. This clearly demonstrates the benefits of combing visual and textual data.
Our proposed techniques perform the best as they improves the accuracy by 8\% and 2\% in comparison with using single modality and early/late fusion, respectively. Among the proposed techniques, common space fusion has higher accuracy. However, the difference is small as it is only 0.6\%.


\begin{table}[t]
\centering
\caption{Performance of different fusion models and two baselines on two datasets}
\label{tbl:exp_reddit}
\begin{tabular}{@{}lcccccc@{}}
\toprule
\textbf{Model}                       & \multicolumn{3}{c}{\textbf{Reddit}}                                                                 & \multicolumn{3}{c}{\textbf{Flickr}}                                                               \\ \cmidrule(lr){2-4}
\cmidrule(lr){5-7}
                  & \multicolumn{1}{l}{\emph{Accuracy}} & \multicolumn{1}{l}{\emph{F-Macro}} & \multicolumn{1}{l}{\emph{F-Micro}} & \multicolumn{1}{l}{\emph{Accuracy}} & \multicolumn{1}{l}{\emph{F-Macro}} & \multicolumn{1}{l}{\emph{F-Micro}} \\ \midrule
Image-based classifier & 77.48                         & 0.77                         & 0.78                         & 56.87                        & 0.67                        & 0.73                        \\
Text-based classifier  & 78.4                          & 0.77                         & 0.78                         & 88.3                         & 0.85                        & 0.88                        \\ \cmidrule(lr){1-7}
Late fusion            & 83.33                         & 0.82                         & 0.83                         & 91.83                        & 0.87                        & 0.91                        \\
Early fusion           & 84.11                         & 0.81                         & 0.83                         & 92.69                        & 0.89                        & 0.92                        \\ \cmidrule(lr){1-7}
Joint fusion           & 86.29                         & 0.84                         & 0.86                         & 93.01                        & 0.90                        & 0.93                        \\
Common space fusion    & \textbf{86.92}                         & \textbf{0.85}                         & \textbf{0.87}                         & \textbf{93.44}                        & \textbf{0.91}                        & \textbf{0.934}                       \\ \bottomrule
\end{tabular}
\vspace{-15pt}
\end{table}

\subsubsection{Results on the Flickr dataset}
The experimental results for the Flickr dataset is also shown in Table \ref{tbl:exp_reddit}. The dataset is interesting as there is a large discrepancy in accuracy (30\%) between image-based and text-based classifiers. On the contrary, the discrepancy between these classifiers in the Reddit dataset is only 1\%. The reasons for this could be 1) the title and description of an image post from flickr are more descriptive as they tend to be longer than the title of a submission from reddit, 2) the labels from the reddit dataset are more reliable as they are ``labelled'' by at least 100 users in comparison with only 5 for the flickr dataset.
Similar to the reddit dataset, all the fusion approaches have better accuracy than unimodal classifiers and our proposed approaches outperform the rest with the common space fusion has better accuracy.

The experimental results on two datasets with different characteristics show that combining textual and visual data can improve the accuracy of classification significantly. In addition, our proposed techniques are robust with different setting as they always achieve the highest accuracy throughout two datasets.


\subsubsection{Handling the absence of one modality}
The goal of joint and common space fusion is not only able to achieve higher accuracy by combining visual and textual signals but also able to provide correct classification when only either text or image is available. In this section, we analyze the performance of our proposed approaches when only one modality is used as input. Table~\ref{tbl:exp_single} shows the experimental results for two different datasets, we also replicate the results with the image and text-only classifiers for comparison purposes. The results clearly demonstrate the benefits of creating a common space as the difference in accuracy between using only image or text as input for common space fusion is significantly lower than that of joint fusion. For the \emph{reddit} dataset, the difference is only 0.8\% for common space fusion while it is 34\% for joint fusion.
The discrepancy for common space fusion is lower as it considers image and text vector of a post as equally important. As a result, using either image or text to classify, we get similar accuracy. On the other hand, joint fusion considers textual information as more important. This makes the classification using text significantly better than using only image. Another key finding is that using only image or text as input, the common space fusion has better result on the flickr dataset than single-input classifier. We achieve a gain of 13\% for image and 0.8\% for text. As the common space contains information from two modalities, even when we use only one modality as input, we can also leverage information from other modality. However, by enforcing a common space from two modalities, we may lose information if one modality is less informative than the other. This happens with the reddit dataset where the titles are shorter and less descriptive than the descriptions of flickr images.

\begin{table}[t]
\centering
\caption{Accuracies of joint and common space fusion when only one modality is available}
\label{tbl:exp_single}
\footnotesize
\begin{tabular}{@{}lllll@{}}
\toprule
                        & \multicolumn{2}{c}{\bf reddit}                           & \multicolumn{2}{c}{\bf flickr}                           \\ \cmidrule(l){2-3} \cmidrule(l){4-5}
                        & \multicolumn{1}{c}{\bf image} & \multicolumn{1}{c}{\bf text} & \multicolumn{1}{c}{\bf image} & \multicolumn{1}{c}{\bf text} \\ \midrule
single-input classifier & 77.48                     & 78.4                     & 56.869                    & 88.3                     \\
joint fusion            & 35.83                     & 69.47                    & 29.46                     & 85.81                    \\
common space fusion     & 71.03                     & 71.96                    & 69.03                     & 89.14                    \\ \bottomrule
\end{tabular}
\vspace*{-15pt}
\end{table}

\subsection{Qualtitative analysis}

\subsubsection{Results with multimodal input}
We also compare and analyze the results to find out in which case our proposed approaches, image-only or text-only approach is better. Table \ref{tbl:full} shows some noteworthy examples. The first two examples demonstrate the benefits of creating the common space as it allows the classifier to put more weights on the visual signal. As the image descriptors of these examples are sarcasm, focusing only on the text give incorrect predictions. For instance, the image descriptor contains the word ``present'' which is a strong indicator for the emotion \emph{joy}. However, the image shows that the present is a creepy doll in a closet, which expresses the emotion \emph{fear}.
Only by taking into account both the image and the text, a classifier can make a correct prediction i.e. this is the advantage of fusion approaches.
The next two examples shows that when a signal is stronger than the other, the fusion approaches will make the same prediction as the stronger signal. In very rare cases as in the fifth example, the stronger signal is an incorrect one, which makes the fusion's prediction incorrect.

\renewcommand\arraystretch{0.5}
\begin{table}[H]
\vspace{-25pt}
\centering
\setlength{\extrarowheight}{1em}
\caption{Qualitative results when both image and text are available}
\label{tbl:full}
\footnotesize
\scalebox{0.95}{
\centering
\begin{tabular}{|l|c|c|c|c|c|}
\hline
\textbf{Image} &\includegraphics[scale=0.2,valign=c]{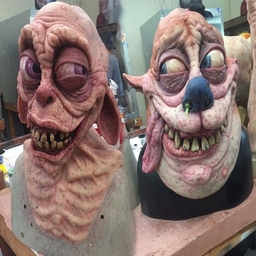} & \includegraphics[scale=0.2,valign=c]{images/2v3t29.jpg} & \includegraphics[scale=0.2,valign=c]{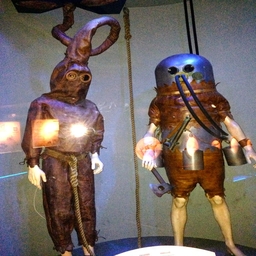} & \includegraphics[scale=0.2,valign=c]{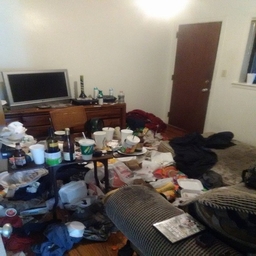} & \includegraphics[scale=0.2,valign=c]{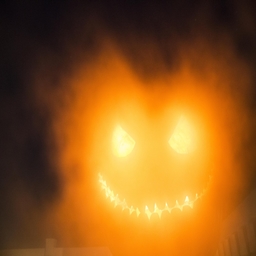}\\
\hline
 \textbf{Image descriptor}&\parbox[t]{2cm}{Happy Happy Joy Joy!!!} & \parbox[t]{2cm}{Today i'm moving, so I decided to leave a little present for the next tenants.} & \parbox[t]{2cm}{Early diving equipment} & \parbox[t]{2cm}{but john , why do n ' t you like your roommate ? }& \parbox[t]{2cm}{I took this long exposure shot of a pumpkin as my lense unknowingly fogged up}\\
 \hline
\textbf{Ground truth} & fear & fear  & fear & anger &fear\\ 
\hline
\textbf{Text classifier}& joy & joy  & disgust & anger &fear\\
\hline
\textbf{Image classifier} & fear & fear& fear & fear  &joy\\
\hline
\textbf{Joint fusion} & joy & joy & fear & anger&joy\\
\hline
\textbf{\parbox[t]{2.5cm}{Common space}} & fear & fear & fear & anger &joy\\
\hline
\end{tabular}
}
\vspace{-15pt}
\end{table}

\subsubsection{Results with the absence of one modality}
Table \ref{tbl:single} shows several examples where either image or text is used as input to our fusion approaches. As one type of input is missing, single-modality fusion approaches may misclassify. This is illustrated in the first two example where unimodal classifiers classify incorrectly while fusion approaches give correct results. In the first example, the presence of the words ``wife'',``baby'' usually indicates the emotion \emph{joy} while the image indicates the emotion \emph{fear}. In the next two examples, common space fusion makes correct predictions while other approaches fail. The text from the third example contains the words ``wake up'', ``morning'', which indicates the emotion \emph{joy}. This makes joint fusion misclassify as it considers textual signal as more important. On the other hand, common space fusion is able to balance between textual and visual signals and come up with correct predictions. The fifth example shows a case where the image-only common space fusion is even better than the image classifier. While the image classifier predicts the emotion \emph{fear} due to the presence of the color red and people's skin, the image actually expresses the emotion \emph{contentment}, which is correctly classified by the image-only common space fusion.


\begin{table}[H]
\centering
\setlength{\extrarowheight}{1em}
\caption{Qualitative results when either image or text is available}
\label{tbl:single}
\footnotesize
\scalebox{0.95}{
\centering
\begin{tabular}{|l|c|c|c|c|c|}
\hline
\textbf{Image} &\includegraphics[scale=0.2,valign=c]{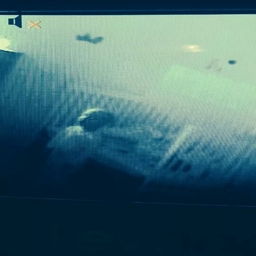} & \includegraphics[scale=0.2,valign=c]{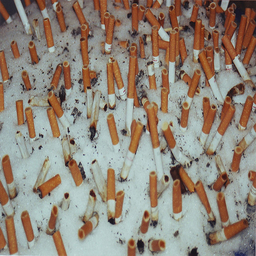} & \includegraphics[scale=0.2,valign=c]{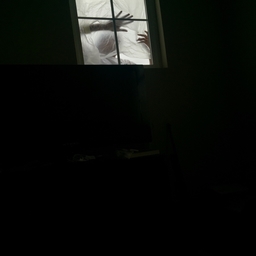} & \includegraphics[scale=0.2,valign=c]{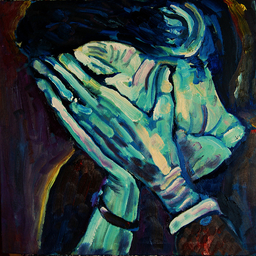}
& \includegraphics[scale=0.2,valign=c]{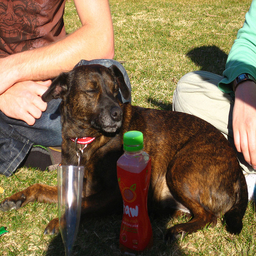}\\
\hline
 \textbf{Image descriptor}&\parbox[t]{2cm}{My wife heard her name being whispered [...] she looked at the baby monitor to find our son like this.} & \parbox[t]{2cm}{New Year's Resolutions Taken on my way home for christmas 2005.} & \parbox[t]{2cm}{I live on the second story and a painter made me almost crap my pants when i woke up this morning} & \parbox[t]{2cm}{A Long Way from Home Oil on canvas by Michael Kent} & \parbox[t]{2cm}{Zelda at the height of contentment. Lily's going away picnic at Ballast Point Park in Balmain.}\\
 \hline
\textbf{Ground truth} & fear & disgust  & fear & \bf sadness &\bf contentment\\
\hline
\textbf{Text classifier}& \textbf{anger} & \textbf{contentment}  & \bf joy & contentment & contentment \\
\hline
\textbf{Image classifier} & \textbf{fear} & \textbf{disgust}&  \bf fear & fear  & \bf disgust\\
\hline
\textbf{Joint fusion} & \bf fear & disgust &  joy &  contentment & contentment\\
\hline
\textbf{Joint fusion - text} & joy & contentment &   joy & contentment &contentment\\
\hline
\textbf{Joint fusion - image} & disgust & amusement &  fear & amusement & amusement\\
\hline
\textbf{\parbox[t]{2.5cm}{Common space}} & \textbf{fear} & \textbf{disgust} & \bf fear & \bf sadness & contentment\\
\hline
\textbf{\parbox[t]{2.5cm}{Common space - text}} & joy & contentment & \bf fear & \bf sadness & contentment\\
\hline
\textbf{\parbox[t]{2.5cm}{Common space - image}} & fear & \textbf{disgust} & \bf fear & contentment & \bf contentment\\
\hline
\end{tabular}
}
\vspace{-10pt}
\end{table}

\section{Conclusions}
\label{sec:conclusion}
In this paper, we propose simple models that are able to combine information from different modalities to analyze social media. Our models are robust with different types of input as they can handle the missing of some modalities. In addition, we show that our models, despite being simple, can achieve high accuracy on the emotion classification application. In order to showcase our models, we also constructed two multimodal datasets which allow us to test our approaches in different settings. Future research directions will go towards analyzing the performance of our models in problems that involve other modalities such as structured data~\cite{nguyen2015result} and user-annotated data~\cite{hung2015minimizing} and other applications beside emotion classification.

\sstitle{Acknowledgement} We gratefully acknowledge the support of NVIDIA Corporation with the donation of the Titan X GPU used for this research. We would also like to thank the IT support team for their help.



\bibliographystyle{splncs03}





\end{document}